%% file: IEEE-conference-template-062824.tex
\documentclass[letterpaper, 10pt, conference]{ieeeconf}
\IEEEoverridecommandlockouts 

\usepackage{cite}
\usepackage{amsmath,amssymb,amsfonts}
\usepackage{algorithmic}
\usepackage{graphicx}
\usepackage{textcomp}
\usepackage{xcolor}
\usepackage{duckuments} 
\usepackage{tikzducks}
\usepackage{graphicx}
\usepackage{caption}
\usepackage{subcaption}
\usepackage{pgfplots}
\usepackage[margin=2cm]{geometry}
\usepackage{array, booktabs, siunitx}
\usepackage{tabularx}
\usepackage{dblfloatfix}
\usepgfplotslibrary{groupplots}
\usepgfplotslibrary{fillbetween}
\pgfplotsset{compat=1.18}
\def\BibTeX{{\rm B\kern-.05em{\sc i\kern-.025em b}\kern-.08em
    T\kern-.1667em\lower.7ex\hbox{E}\kern-.125emX}}

\usetikzlibrary{external}
\usepackage{textcomp}
\usepackage{lipsum}
\newcommand\copyrighttext{%
	\footnotesize \textcopyright 2026 IEEE.  Personal use of this material is permitted.  Permission from IEEE must be obtained for all other uses, in any current or future media, including reprinting/republishing this material for advertising or promotional purposes, creating new collective works, for resale or redistribution to servers or lists, or reuse of any copyrighted component of this work in other works.
}
\newcommand\copyrightnotice{%
	\tikzset{external/export=false}
	\begin{tikzpicture}[remember picture,overlay]
	\node[anchor=south,yshift=10pt, xshift=10pt] at (current page.south) {\fbox{\parbox{\dimexpr\textwidth-\fboxsep-\fboxrule\relax}{\copyrighttext}}};
	\end{tikzpicture}%
	\tikzset{external/export=true}
}

\begin{document}

\title{Belief-Space Residual Risk for Automated Driving under Localization Uncertainty}

\author{Nijinshan Karunainayagam$^{1,*}$, Nils Gehrke$^{1}$, Frank Diermeyer$^{1}$
\thanks{$^{1}$Authors are with the Institute of Automotive Technology at the Technical University of Munich (TUM), DE-85748 Garching, Germany. $^{*}$Corresponding author: {\tt\small nijinshan.karunainayagam@tum.de}}}%

\maketitle
\copyrightnotice

\begin{abstract}
Residual risk metrics have recently been introduced to assess the safety implications of automated driving systems. Existing approaches typically assume a deterministic ego pose and concentrate mainly on perception errors related to surrounding objects and latency effects. In practice, however, automated vehicles operate under considerable localization uncertainty, especially in complex urban settings and in adverse weather conditions. This work extends the spatial residual risk formulation to the belief space by explicitly modeling ego pose uncertainty as a Gaussian distribution. Residual risk is reformulated as the expected degradation-induced risk over the ego pose belief distribution. Within a particle-based risk estimation framework, localization uncertainty is incorporated into the computation of collision probabilities through covariance fusion of ego and object uncertainties.
\end{abstract}


\input{sections/01_Introduction.tex}
\input{sections/02_Related_Work}
\input{sections/03_Methodology}

\input{sections/04_ExperimentalSetup}

\input{sections/05_Results}
\input{sections/06_Conclusion}

\bibliographystyle{IEEEtran}
{\small
\bibliography{bib}}

\end{document}

%% file: sections/01_Introduction.tex
\section{\uppercase{Introduction}}
\label{sec:introduction}

Ensuring overall safety remains one of the central challenges in bringing Automated Vehicles (AVs) into large-scale production. The high level of system complexity required to operate reliably under environmental uncertainty makes it difficult to guarantee safe behavior across all operational conditions. To address this challenge, modern AV architectures implement multiple safety mechanisms, including continuous risk assessment processes that consider both external scene information and internal system states.

\subsection{Motivation}
Driving risk is commonly defined as the combination of the probability of occurrence of a situation and the expected severity of its outcome \cite{iso26262}. Risk increases when critical situations emerge. Such situations may be caused by external factors like environmental conditions or other road user's behavior \cite{Guo2020}. The resulting system degradation influences the AV's decision making process and may lead to behavior that deviates from the nominal system response. Even if the environment remains unchanged, a degraded system may select suboptimal behaviors, thereby increasing collision risk.

To quantify this degradation-induced safety loss, residual risk is introduced. Residual risk compares the risk of a degraded system to that of a non-degraded reference system under identical environmental conditions. Existing residual risk frameworks typically model degradation through perturbations in perceived object state or through latency in sensing and processing. These approaches focus on errors in detection, tracking and prediction of surrounding traffic participants. However, the ego vehicle’s own pose is generally treated as deterministic and ideal.

\begin{figure}[!t]
    \centering
    \includegraphics[width=\linewidth]{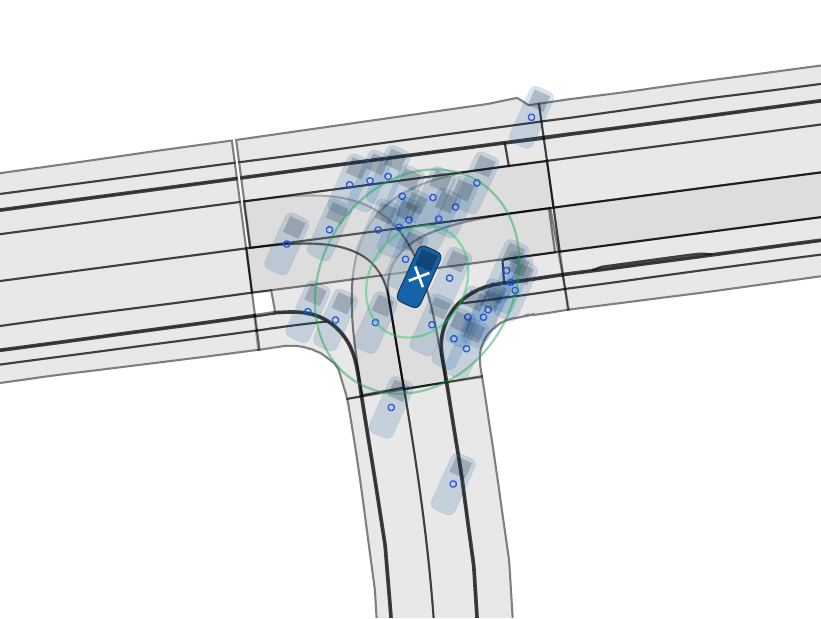}

    \caption{Visualization of ego localization uncertainty on a T Section of the TUMDOT \cite{Kutsch2024} dataset}
    \label{fig:catchy_figure}
\end{figure}

In practical robotics systems, this assumption does not hold. Ego pose estimates provided by Gaussian localization algorithms are inherently probabilistic and are commonly represented as a state estimate with associated covariance \cite{Li_Xifeng_State_2013}. Especially in GNSS-degraded urban environments or under limited sensor visibility, localization uncertainty can grow significantly. Elevated ego localization uncertainty directly affects the relative geometry between the ego vehicle and surrounding objects. Even small positional uncertainties can lead to critical situations, especially when driving in narrow urban streets with parked vehicles or in intersections (see Fig. \ref{fig:catchy_figure}). This can alter collision probabilities, invalidate safety margins, and potentially violate motion planning constraints.

Despite its practical relevance, ego localization uncertainty has not been explicitly integrated into residual risk formulations. This work extends the residual risk concept into belief space by incorporating ego pose uncertainty into the risk computation framework.

\subsection{Contribution}
Since most risk assessment (RA) methods consider the AV's position as deterministic, this work investigates the influence of positional uncertainty of the ego vehicle on the driving risk. To tackle this, an existing spatial residual risk definition is extended and adapted by the influence of positional uncertainty of the ego vehicle. This is analyzed and quantified for different localization uncertainty modes. The experiments are conducted on real traffic data from the TUMDOT-MUC dataset \cite{Kutsch2024}. The main contributions of this work are:
\begin{itemize}
    \item \textbf{Belief-Space Extension of Spatial Residual Risk} An existing spatial residual risk formulation \cite{Gehrke2025} is extended. This is done by relaxing the deterministic ego pose assumption and introducing a probabilistic belief-space representation of the ego state. The residual risk is reformulated as an expectation over ego pose uncertainty.
    \item \textbf{Quantitative Analysis of Localization-Induced Risk Amplification} A systematic analysis is conducted how increasing ego pose covariance influences residual risk under different degradation modes and demonstrate nonlinear risk amplification effects in realistic urban traffic scenarios
    \item \textbf{Online-Capable Implementation and Evaluation on Real Traffic Data} The proposed extension is integrated into a particle-based residual risk framework and evaluated on the TUMDOT-MUC dataset, demonstrating feasibility for online safety assessment.
\end{itemize}

%% file: sections/02_Related_Work.tex
\section{\uppercase{Related Work}}
Safety violations typically arise from system failures or defects, which may be triggered or amplified by uncertainties in dynamic and critical driving situations \cite{Lu2025RiskAI} \cite{Gehrke2025}. In this section, the state of the art in risk assessment methods and approaches that explicitly account for uncertainty are reviewed.

\subsection{Risk Assessment for Automated Vehicles}
According to ISO 26262, safety is defined as the absence of unreasonable risk \cite{iso26262}. Thus, safety can be interpreted as the reduction of risk to an acceptable level, allowing risk to serve as a quantitative indicator of system safety. Risk is commonly defined as the combination of the probability of occurrence of a situation $p$ and the expected severity  $s$ of its consequences:
\begin{equation}
    \label{eq:risk}
    R(situation) = p(situation) \cdot s(situation).
\end{equation}

Risk Assessment (RA) in AVs can be generally distinguished into development-time (offline) and runtime (online) RA \cite{Chia2022}. During AV development, regulatory guidelines prescribe safety analysis procedures. Thus, the automotive standard ISO 26262 \cite{iso26262} requires the execution of Hazard Analysis and Risk Assessment (HARA) to classify hazardous events into Automotive Safety and Integrity Levels (ASIL) based on severity, probability and controllability. Aside of this, further methods to identify failures and hazards are used from a functional safety point of view, such as Failure Mode and Effects Analysis (FMEA)\cite{mikulak2017basics}, Fault Tree Analysis (FTA)\cite{vesely1981fault} or Systems-Theoretic Process Analysis (STPA) \cite{Leveson2021}. Furthermore, standards such as SAE J3016 considers environmental uncertainties affecting the risk of the system \cite{saej3016}. Development-time RA is conservative and mostly scenario-based. It depends on assumptions and cannot capture the combinatorial complexity of real-world interactions. Consequently, not all potential hazards can be identified beforehand.

Therefore, runtime RA is crucial for safety and safe operation during the driving mission. Runtime RA captures the current situation and the system's state to compute driving risk. Due to its real-time capability and uncertainty-awareness, runtime RA complements development-time RA by addressing residual risk. Although no general taxonomy exists, runtime RA can be categorized in rule-based (e.g. \cite{Khayatian2021}), probability-based (e.g. \cite{KATRAKAZAS201961}) and learning-based approaches (e.g. \cite{Strickland2018}). \cite{Lu2025RiskAI} branch RA into an additional equivalent model-based approaches, including methods like Fuzzy Logic, Potential fields or Risk Field Theory. 

When incorporating uncertainties into RA, probabilistic approaches are the most promising methods. While rule-based, model-based and learning-based approaches typically handle uncertainty implicitly, probabilistic methods provide a more sophisticated mathematical treatment of uncertainty. The limiting factors of probabilistic methods are increased computational and modeling complexity.

\subsection{Risk Assessment under Uncertainty}

Based on this definition, risk assessment methods aim to estimate either the likelihood of critical scenarios, the associated severity, or both. Since the likelihood of hazardous events and the corresponding wrong or unsafe reaction quantifies uncertain future occurrences \cite{Torngren2018}, uncertainty can be seen as a structural element of risk assessment. 

Uncertainty can be differed into aleatoric and epistemic uncertainty, commonly caused by the behavior of other traffic participants, occlusions, sensor noise or environmental conditions like adverse weather \cite{Araujo2024,SHETTY2021,Chougule2024}. These affects the system and its modules directly by causing degradation in localization, perception and decision making. Therefore the need of incorporating such uncertainties into risk assessment is crucial for ensuring AV safety.

One research direction focuses on computing the probability of collision between uncertain agents. Early probabilistic collision checking formulation model collision avoidance as a chance constraint over uncertain robot and obstacle states, explicitly integrating joint state distributions into collision evaluation \cite{Toit2011}. 
In contrast, \cite{Hakobyan2019} tackles the coherency and convexity of chance constraints by introducing conditional value-at-risk (CVaR) to regulate motion safety under stochastic obstacle motion.  

To Improve computational tractability, several works derive analytic or semi-analytic probability of collision approximations. In \cite{Tolksdorf24} efficient collision probability estimation using multi-circular shape approximations are used, enabling real-time evaluation while bounding approximation errors. Similarly, \cite{Kaufeld2025} introduce semi-analysis formulations based on spatial overlap and stochastic boundary crossing probabilities, incorporating full pose uncertainty while maintaining real-time performance. 

Another method embeds uncertainty directly into motion planning through chance constraints, ensuring that the probability of constraint violation remains below a predefined threshold \cite{Jasour2019}. They introduce risk contour maps, which encode spatial regions according to admissible collision probabilities and transform stochastic safety requirements into deterministic planning constraints.

In \cite{Gehrke2025} the uncertainty caused by the misperception of environmental objects and traffic participants is considered in the risk assessment by including it into a residual risk formulation. The formulation compares the potential crash risk of a degraded and non-degraded system.

\subsection{Residual Risk Formulation}

According to \cite{Gehrke2025}, the residual risk can be defined considering the baseline risk of a system compared to a risk induced by a unsuitable response to an opponent's trajectory caused by a degraded system. Considering the ego trajectory $T_{1}$ and a set of possible opponent trajectories $T_{2}$, the overall risk is defined as 

\begin{equation}
    \label{eq:risk_formulation}
    R(\mathcal{T}_{1}\otimes \mathcal{T}_{2}) = \sum_{\mathcal{T}_{2}}p(T_{2}) \cdot s(T_{1}|T_{2}, T_{2})
\end{equation}

where $p(T_{2})$ denotes the probability of opponent trajectory $T_{2}$ and $s$ depicts the severity composed of collision probability and and impact dynamics.

Under degraded perception a degraded risk 
\begin{equation}
    \hat{R}(\mathcal{T}_{1}\otimes \mathcal{T}_{2}) = \sum_{\mathcal{T}_{2}}p(T_{2}) \cdot s(\hat{T_{1}}|T_{2}, T_{2})
\end{equation}

can be determined by considering a suboptimal trajectory $\hat{T_{1}}$. The resulting difference in risk caused by degradation defines the residual risk shown in \eqref{eq:residual_risk}.
\begin{equation}
    \label{eq:residual_risk}
    R_{residual} =\hat{R} - R, \qquad with \quad \hat{R} \geq R
\end{equation}

In \cite{Gehrke2025} perceptional degradation is considered in particular, characterized by latency $\theta$ and the opponent's positional error $\epsilon_{x}$ and velocity error $\epsilon_{v}$ resulting in a formal residual risk definition of 
\begin{equation}
     R_{residual} = \hat{R}(\mathcal{T}_{1}\otimes \mathcal{T}_{2}, \theta, \epsilon_{x}, \epsilon_{v})-R(\mathcal{T}_{1}\otimes \mathcal{T}_{2}).
\end{equation}
This allows a sophisticated consideration of the effects of degradation in detection, tracking or prediction. However, the ego position is always considered as ideal...

Some of the references consider localization uncertainty of the ego vehicle in their approach. However, none of these works incorporates the uncertainty estimation within a residual risk formulation. This gap is investigated and filled by this work.

%% file: sections/03_Methodology.tex
\begin{figure}[!t]
    \centering
    \includegraphics[width=\linewidth]{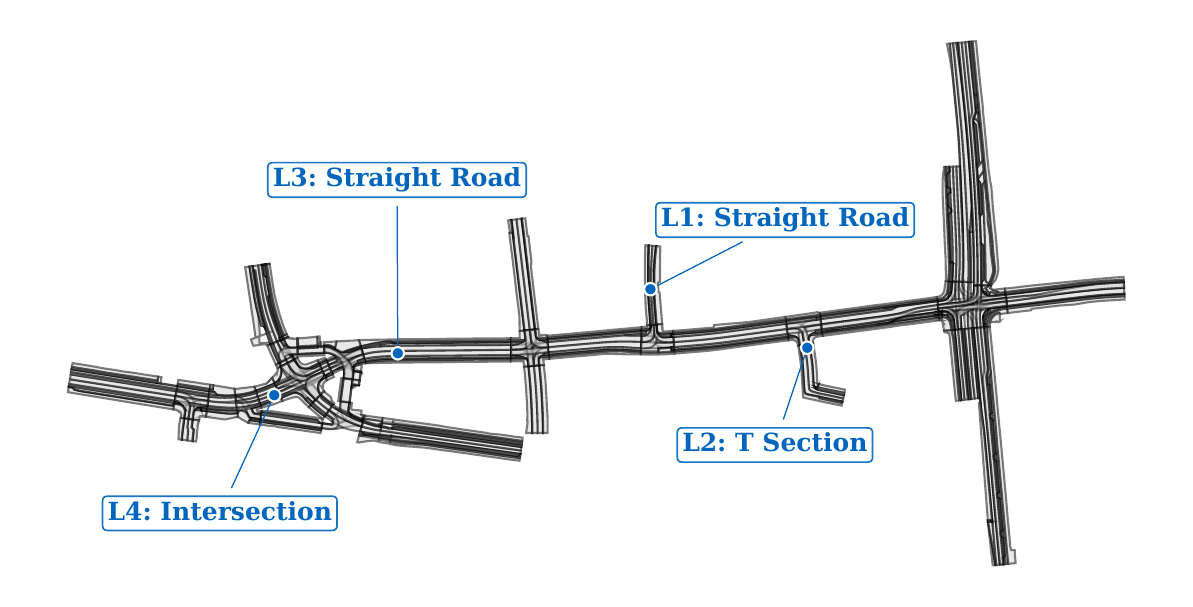}
    \caption{Visualization of the TUMDOT-Dataset with the selected scenario locations marked}
    \label{fig:TUMDOT-Locations}
\end{figure}

\section{\uppercase{Methodology}}
\label{sec:methodology}
\begin{figure*}[!tbp]
    \centering
    \begin{subfigure}[t]{0.48\textwidth}
        \centering
        \includegraphics[width=\linewidth]{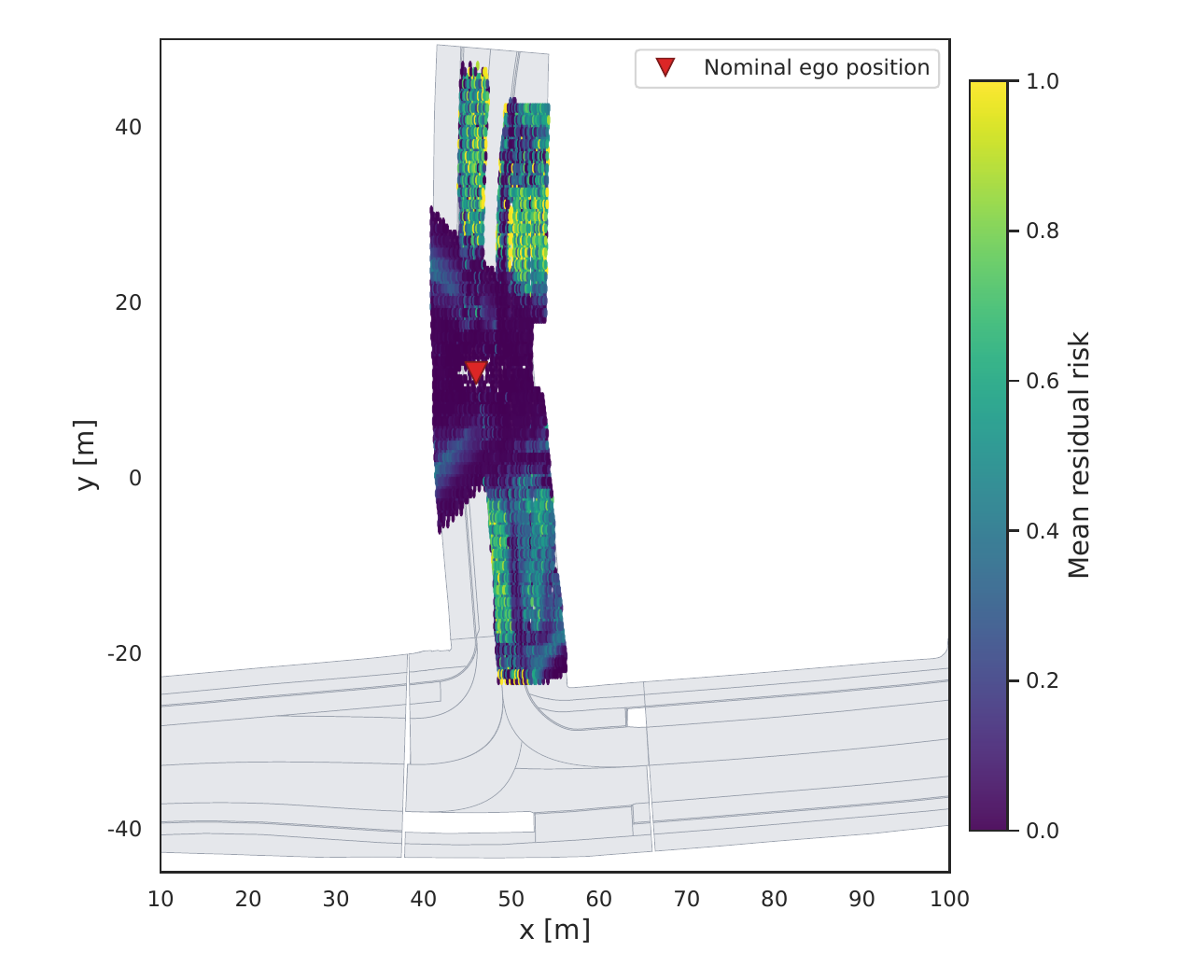}
        \caption{L1: Straight Road}
        \label{fig:APE_over_opening_angle}
    \end{subfigure}
    \hfill 
    \begin{subfigure}[t]{0.48\textwidth}
        \centering
        \includegraphics[width=\linewidth]{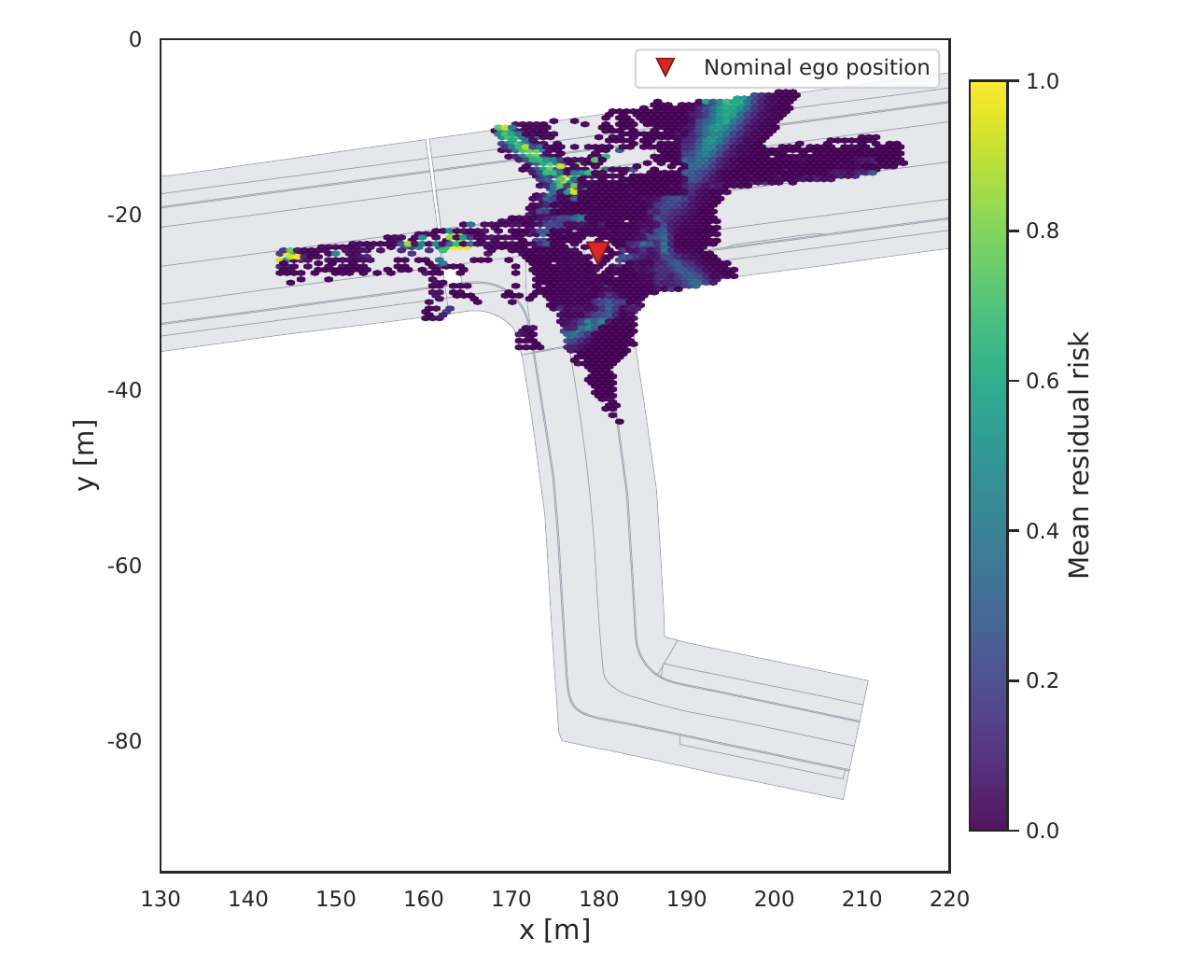}
        \caption{Location L2: T-Section}
        \label{fig:RPE_over_opening_angle}
    \end{subfigure}
    \hfill 
    \begin{subfigure}[t]{0.45\textwidth}
        \centering
        \includegraphics[width=\linewidth]{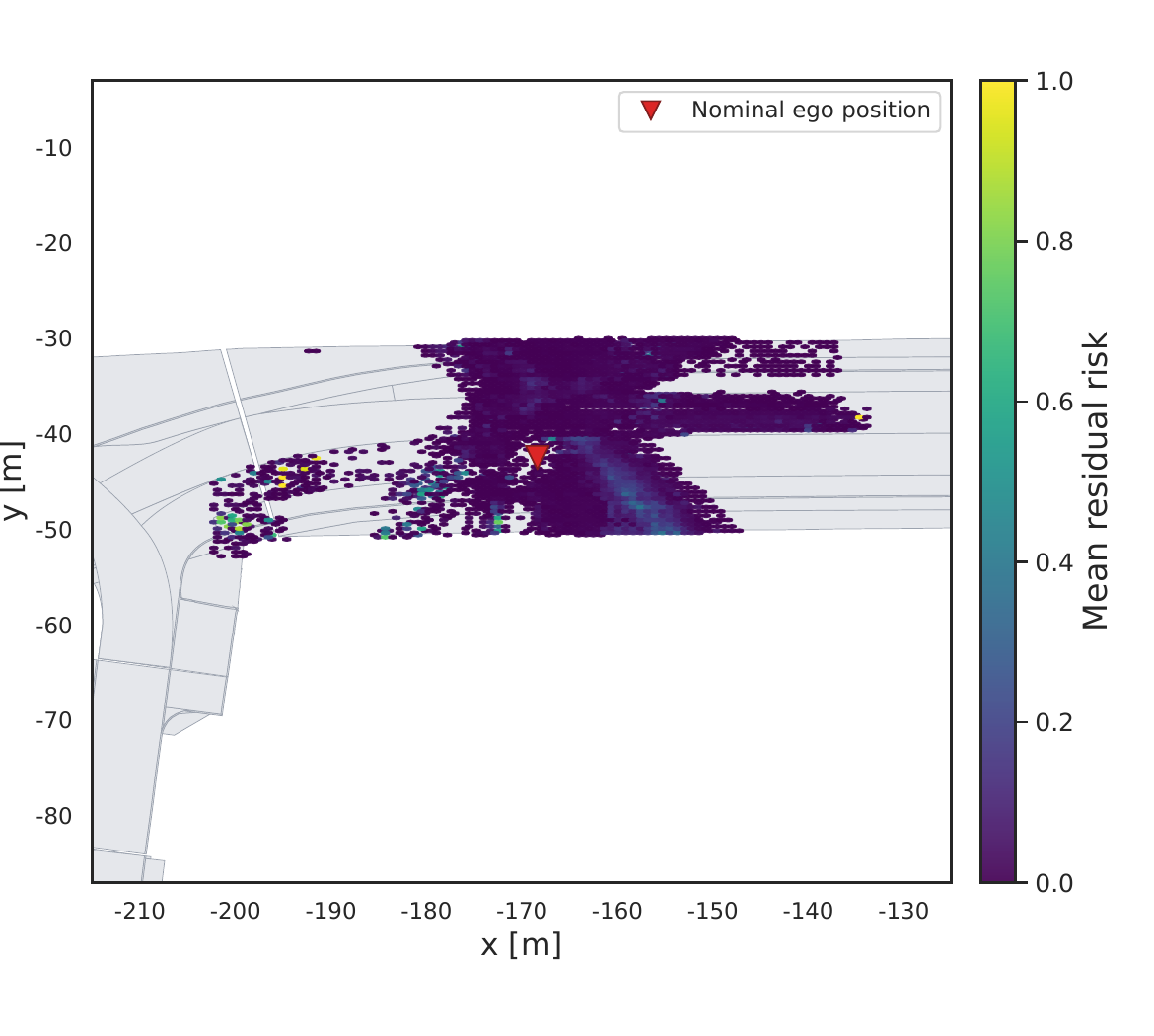}
        \caption{Location L3: Straight Road }
        \label{fig:RPE_over_opening_angle}
    \end{subfigure}
    \hfill 
    \begin{subfigure}[t]{0.45\textwidth}
        \centering
        \includegraphics[width=\linewidth]{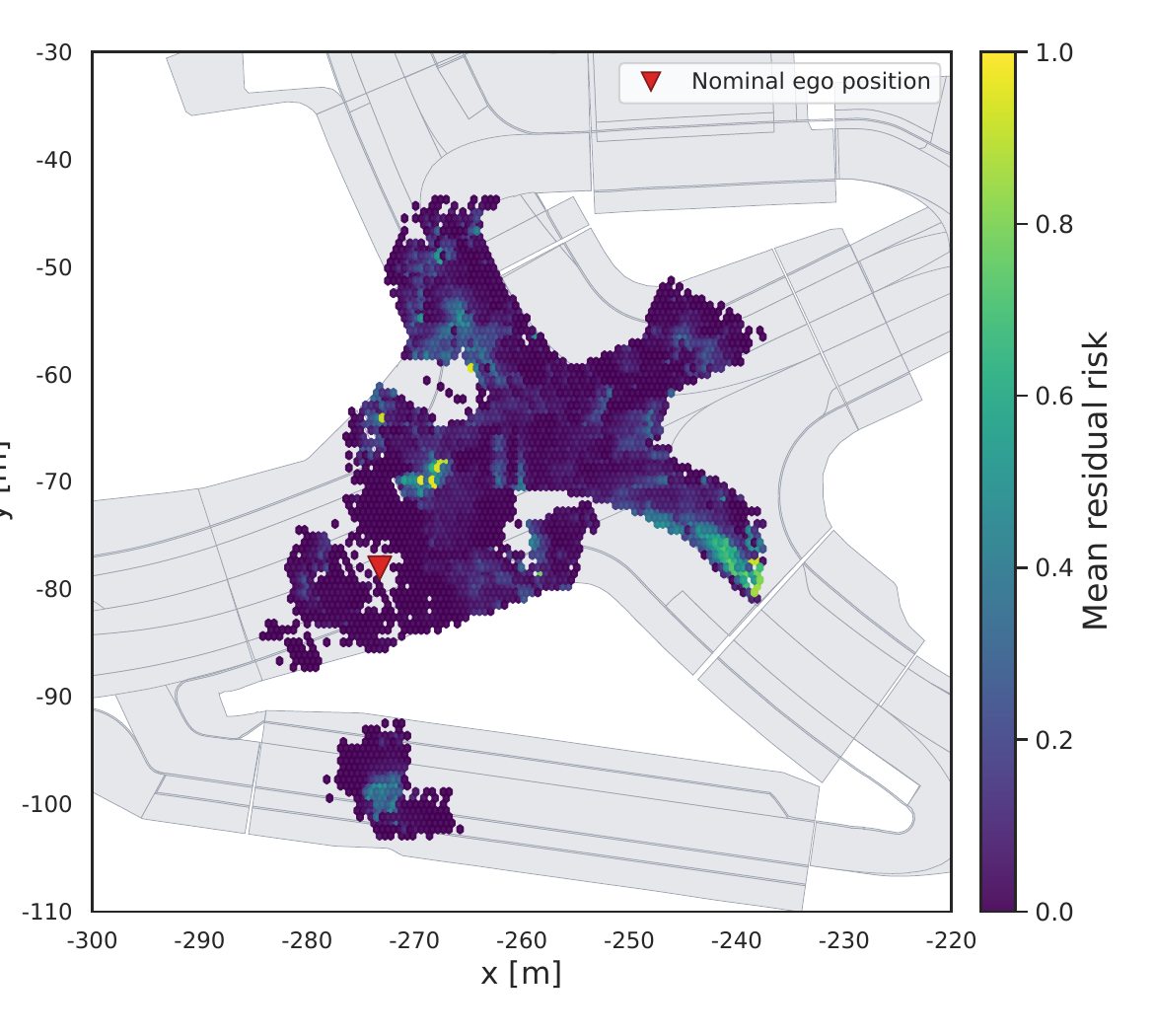}
        \caption{Location L4: Complex Intersection}
        \label{fig:rrss_heatmaps}
    \end{subfigure}
    
    \caption{Spatial heat maps of the mean residual risk of the selected scenarios. The nominal ego positions are each marked for each heat map}
    \label{fig:rr_heatmaps}
\end{figure*}

In the following, a methodology to incorporate localization uncertainty into the determination of spatial residual risk is presented. Building on \cite{Gehrke2025}, the residual risk definition is extended by a gaussian distribution to depict the uncertainty of the ego vehicle. For this, a two dimensional uncertainty formulation is used, consisting of a longitudinal uncertainty $\sigma_{x}$, a lateral uncertainty $\sigma_{y}$. Since yaw uncertainty $\sigma_{yaw}$ will mostly result in lateral uncertainty, it is not considered for this study..

To factor in localization uncertainty in the residual risk computation, the true ego state is modeled as a Gaussian random variable
\begin{equation}
    \mathbf{x}_{e} \sim \mathcal{N}(\mathbf{\hat{x}}_{e},\Sigma_{e})
\end{equation}
where $\hat{\mathbf{x}}_{e}$ is the estimated ego pose and $\Sigma_{ego}$ depicts the localization covariance matrix. This introduces epistemic uncertainty into the ego pose.

Since the severity term introduced in \eqref{eq:risk_formulation} depends on the true relative configuration between ego and opponent, risk becomes conditional on the true ego pose: 
\begin{equation}
    R(\mathbf{x}_{e}) = \sum_{\mathcal{T}_{2}}p(T_{2}) \cdot s(T_{1}|T_{2}, T_{2};\mathbf{x}_{e}).
\end{equation}

The degraded belief-space risk is defined as the expectation over ego pose uncertainty, which can be formulated as 
\begin{equation}
    \tilde{R} = \mathbb{E}_{x_{e}}[R(\mathbf{x}_e)] = \int R(\mathbf{x}_{e}) \cdot \mathcal{N}(\mathbf{x}_e;\hat{\mathbf{x}}_e, \Sigma_{e})\,d\mathbf{x}_{e}.
\end{equation}
Thus, degraded risk becomes
\begin{equation}
    \tilde{\hat{R}} = \mathbb{E}_{x_{e}}[\hat{R}(\mathbf{x}_e)]
\end{equation}
so that the belief-space residual risk can be defined as
\begin{equation}
   \tilde{R}_{residual} = \mathbb{E}_{x_{e}} [\hat{R}(\mathbf{x}_{e}) - R(\mathbf{x}_{e})]. 
\end{equation}

\subsection{Collision Probability under Localization Uncertainty}
Collision probability depends on the relative position between the ego vehicle $\mathbf{x}_e$ and the opponent $\mathbf{x}_o$. If the opponent state is modeled as 
\begin{equation}
    \mathbf{x}_o \sim \mathcal{N}(\mathbf{\overline{x}}_o, \Sigma_{o}),
\end{equation}
and assuming independence between ego and opponent uncertainty, the relative state is Gaussian:
\begin{equation}
    \Delta \mathbf{x} \sim \mathcal{N}(\Delta \overline{\mathbf{x}}, \Sigma_{rel}),
\end{equation}
with covariance
\begin{equation}
    \Sigma_{rel} = \Sigma_{o} + \Sigma_{e}
\end{equation}
This covariance fusion follows directly from properties of Gaussian random variables. The collision probability is then computed as 
\begin{equation}
    P_{coll} = \int_{C} \mathcal{N}(\Delta \mathbf{x}; \Delta \overline{\mathbf{x}}, \Sigma_{rel}) d\Delta\mathbf{x}, 
\end{equation}
where $C$ denotes the collision region. Thus, localization uncertainty increases collision probability through covariance inflation in the relative configuration space.

%% file: sections/04_ExperimentalSetup.tex
\section{\uppercase{Experimental Setup}}
To conduct the analysis, a Monte Carlo sampling is used along different selected scenarios within the TUMDOT-dataset. In the following, the experimental setup for this analysis is presented.
\subsection{Scenario Selection}
Different reference locations on the TUMDOT-MUC road network are selected, representing different traffic situations, namely two straight road segments (L1 and L3), a T-section (L2) and a complex intersection (L4) as shown in \ref{fig:TUMDOT-Locations}. This allows the influence of different road geometries on the residual risk. 

\subsection{Monte Carlo Procedure}
To analyze the sensitivity of the residual risk under positional uncertainty, the particle based residual risk estimation proposed in \cite{Gehrke2025} is used. For this, a Monte Carlo estimate of the expected residual risk is computed by repeatedly sampling a uncertain ego position $\sigma_{ego} $. The base degradation scenario uses a fixed temporal latency T, which allows the baseline residual risk to be non-zero, while the opponents' position and velocity errors are set to zero. That ensures the contribution isolation of the ego localization uncertainty solely.

For each reference location and each ego localization uncertainty the expected residual risk is estimated via Monte Carlo sampling. For this, an uncertain ego position is sampled first before the residual risk computation pipeline is executed with this position. If the sampled ego position is off from valid drivable area like in a map gap or outside of vector field coverage no particle couples are created. If such a sample is created, the particular iteration will be stopped and a new sample is generated. For every valid iteration the residual risk value is computed and stored. These steps are repeated until the desired sampling length is reached. After all runs, the expected residual risk is computed from the sample set. At $\sigma_{ego}=0$ a single deterministic evaluation is performed. For almost all non-zero uncertainty levels, the desired sample length is 30. Solely for two runs 15 Monte Carlo samples are defined, due to the the vast number of particles for high uncertainties (rephrase). 

\subsection{Parameterization}
For each run, the ego localization uncertainty is evaluated across seven levels, ranging from perfect localization ($\sigma_{ego}=0$m) to severe uncertainty ($\sigma_{ego}=2$m). For comparison, an additional sweep over temporal latency $T$ is conducted at $\sigma_{ego}=0$, which provides a direct side-by-side comparison between spatial uncertainty and temporal degradation.

The GHR-Parameterization proposed in \cite{Gehrke2025} is reused for this experiment. Three additional parameters are adapted. First, an ego velocity $v_{ego}=30km/h$ is used for all runs, due to the selected urban scenarios. Based on this, the predicted time horizon over which particles are propagated to evaluate collision risk is set to $2s$. Due to computational efficiency the particle spacing is set to $0.5$.

%% file: sections/05_Results.tex
\begin{figure}[t]
    \centering
    \begin{subfigure}[b]{0.48\linewidth}
        \centering
        \includegraphics[width=\linewidth]{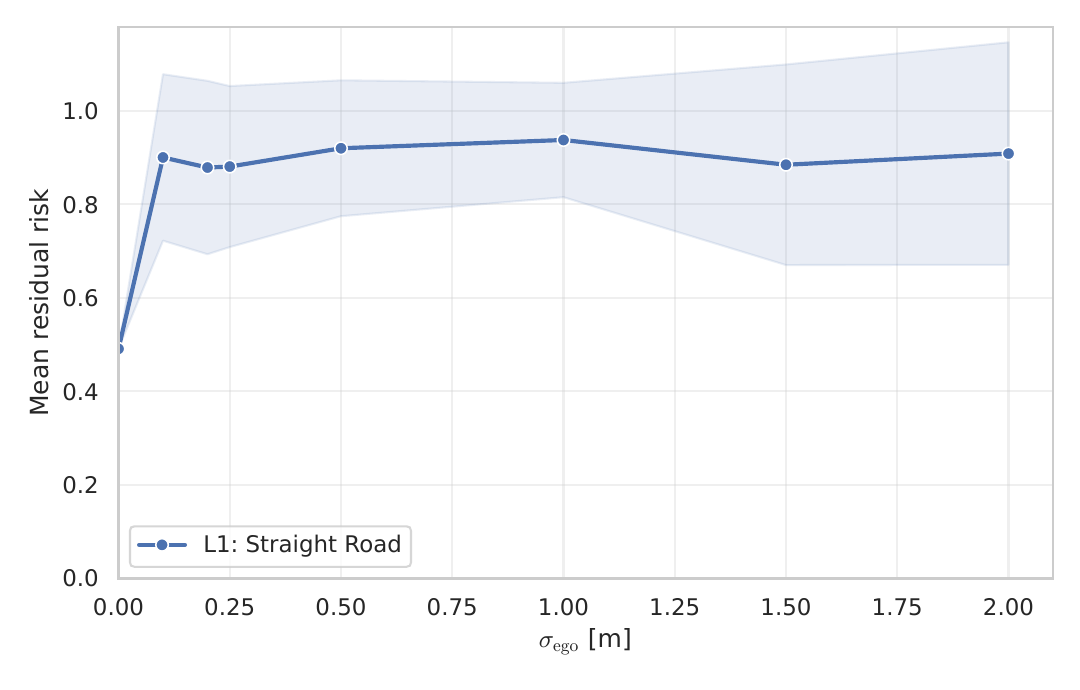}
        \caption{T1: Straight Road}
        \label{fig:crop_full}
    \end{subfigure}
    \hfill
    \begin{subfigure}[b]{0.48\linewidth}
        \centering
        \includegraphics[width=\linewidth]{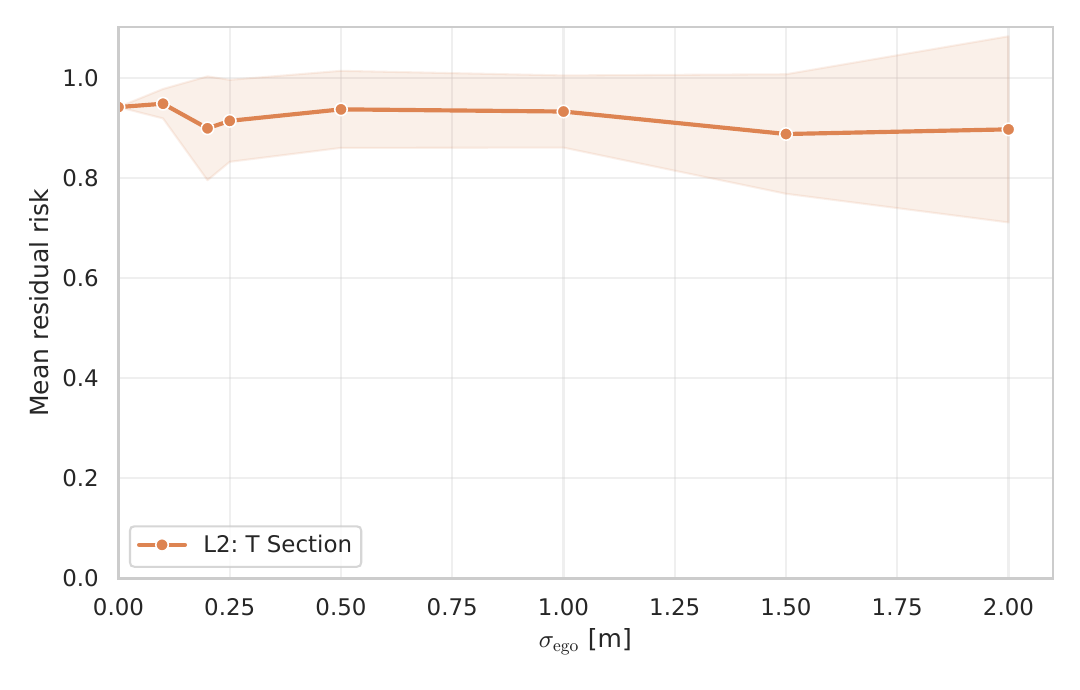}
        \caption{T2: T-Section}
        \label{fig:crop_min45_45}
    \end{subfigure}
    \vspace{0.8em}
    \begin{subfigure}[b]{0.48\linewidth}
        \centering
        \includegraphics[width=\linewidth]{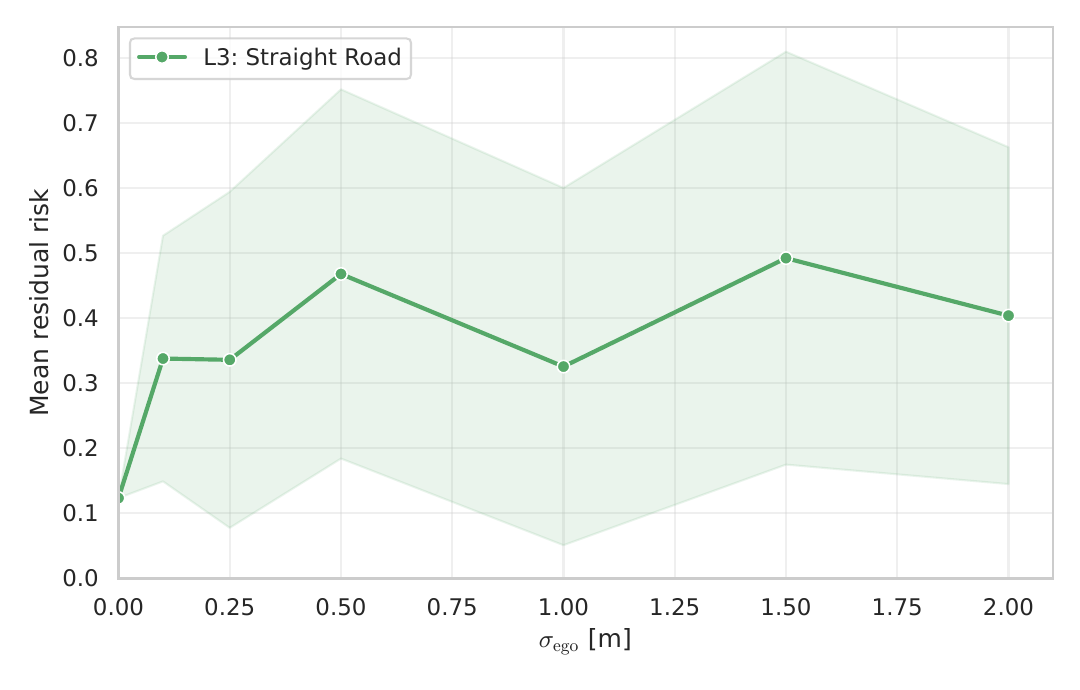}
        \caption{T3: Straight Road}
        \label{fig:crop_200_360}
    \end{subfigure}
    \hfill
    \begin{subfigure}[b]{0.48\linewidth}
        \centering
        \includegraphics[width=\linewidth]{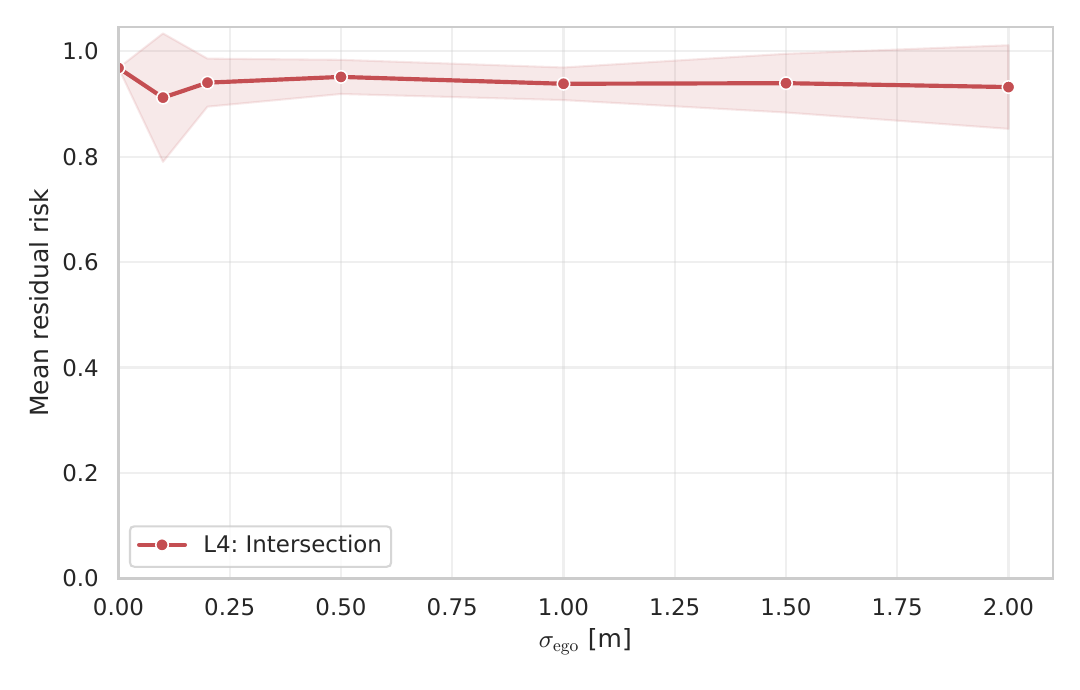}
        \caption{T4: Complex Intersection}
        \label{fig:crop_0_135}
    \end{subfigure}
    \caption{Behavior of the mean residual risk with increasing localization uncertainties for the selected scenarios}
    \label{fig:mean_residual_risk}
\end{figure}

\begin{figure}[t]
    \centering

    \begin{subfigure}[b]{0.48\linewidth}
        \centering
        \includegraphics[width=\linewidth]{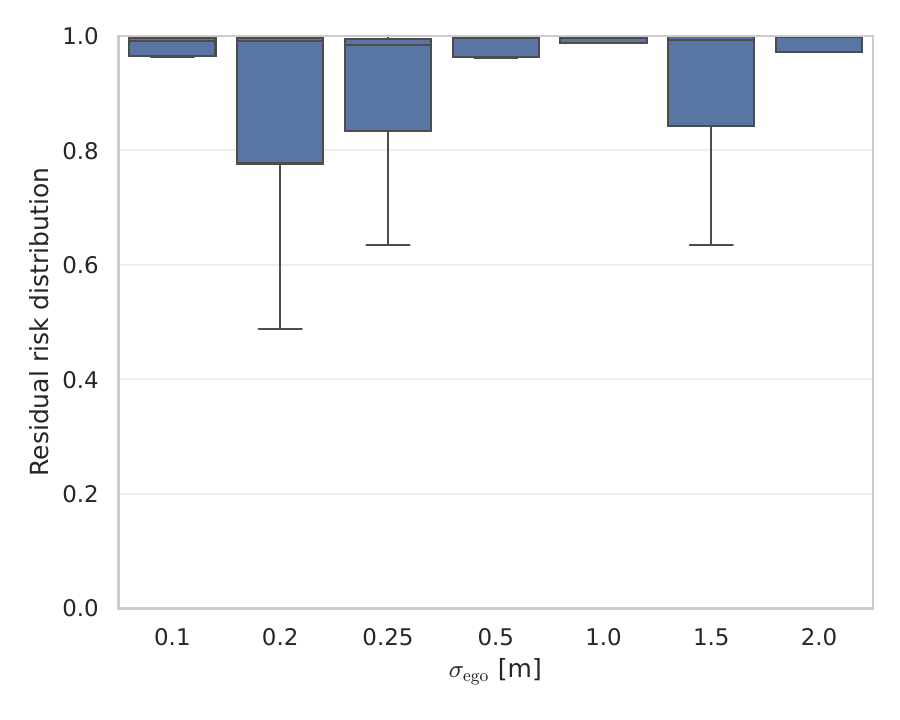}
        \caption{T1: Straight Road}
        \label{fig:crop_full}
    \end{subfigure}
    \hfill
    \begin{subfigure}[b]{0.48\linewidth}
        \centering
        \includegraphics[width=\linewidth]{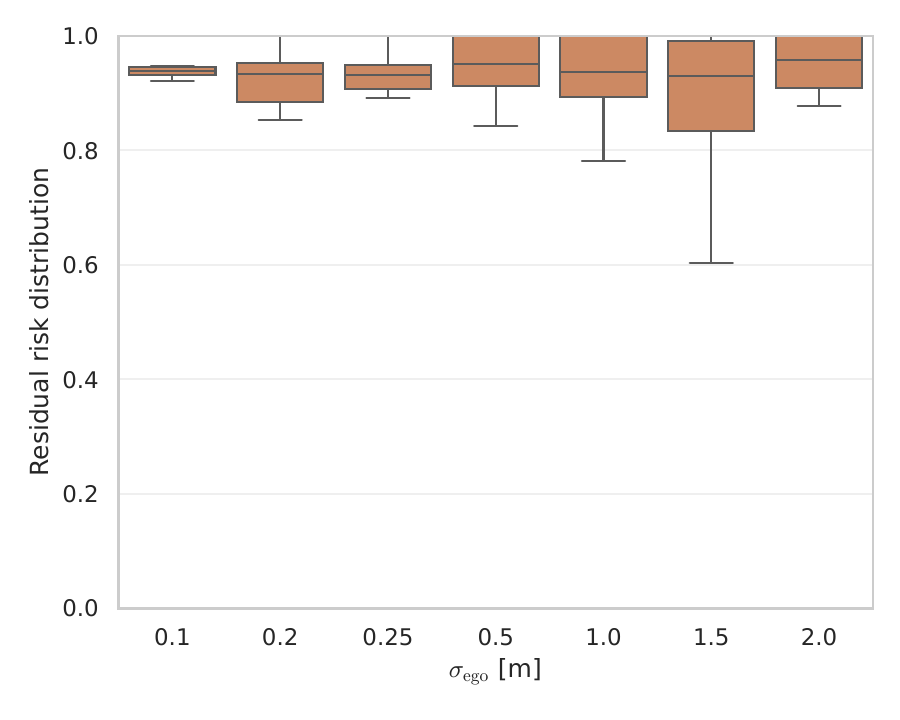}
        \caption{T2: T-Section}
        \label{fig:crop_min45_45}
    \end{subfigure}
    \vspace{0.8em}
    \begin{subfigure}[b]{0.48\linewidth}
        \centering
        \includegraphics[width=\linewidth]{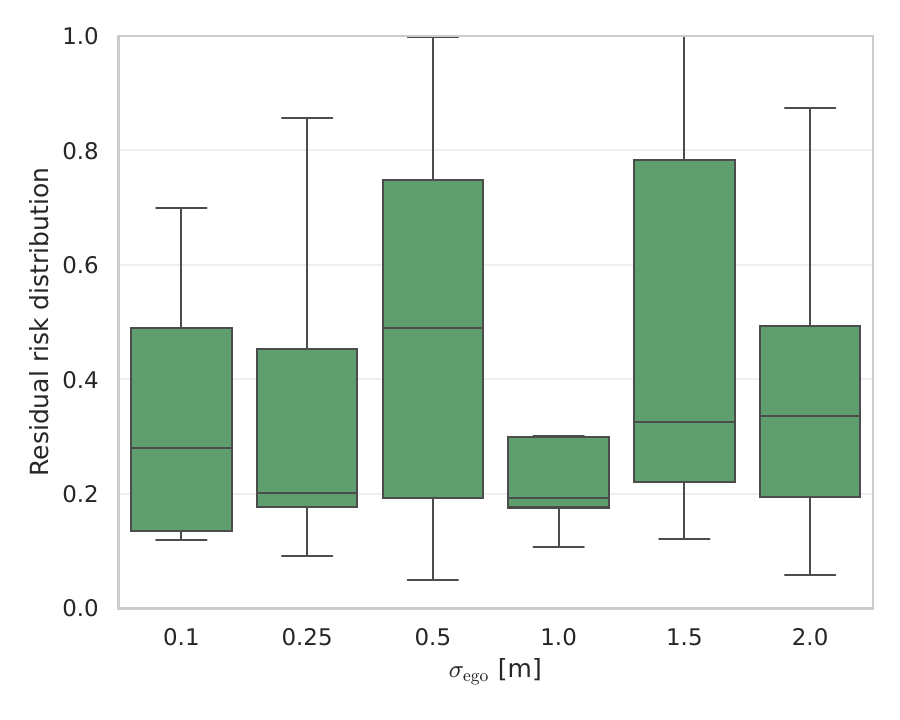}
        \caption{T3: Straight Road}
        \label{fig:crop_200_360}
    \end{subfigure}
    \hfill
    \begin{subfigure}[b]{0.48\linewidth}
        \centering
        \includegraphics[width=\linewidth]{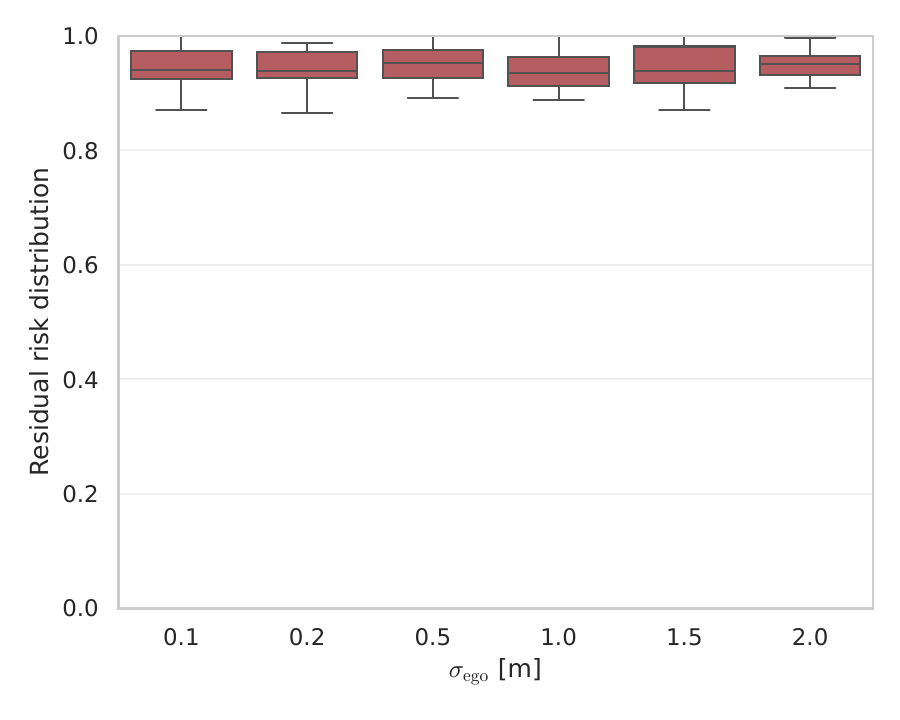}
        \caption{T4: Complex Intersection}
        \label{fig:crop_0_135}
    \end{subfigure}
    \caption{Risk variability over increasing localization uncertainties for the selected scenarios}
    \label{fig:risk_variability}
\end{figure}

\section{\uppercase{Results}}
\label{sec:results}

\subsection{Spatial Residual-Risk Field}
Fig. \ref{fig:rr_heatmaps} illustrates the spatial distribution of the mean residual risk for the selected locations. The heatmaps depict residual risk as a function of ego position under fixed temporal latency. 

\subsubsection{Straight Road (L1 \& L3)} For both straight road scenarios (L1 and L3) the residual risk field varies smoothly along the driving direction. The risk is comparatively low except of several conflict directions. Instead, risk increases gradually along potential conflict directions.  
\subsubsection{T-Section (L2)} Considering the T-Section the residual risk shows a localized high-risk region near the intersection. Strong spatial gradients are visible around this region.
\subsubsection{Complex Intersection (L4)} The complex intersection (L4) shows multiple high risk regions. The spatial topology is irregular with steep gradients and fragmented high risk areas.


\subsection{Baseline Residual Risk}
For deterministic localization, the baseline residual risk shown in table \ref{tab:baselinerr} creates the fundamental reference for further analysis by corresponding to the value of the spatial field at the nominal ego pose. While the straight road scenarios exhibit low to moderate baseline residual risk, the T-Section and complex intersection scenarios show high baseline residual risk. This indicates that the ego pose is already located in a critical configuration. 
\begin{table}[ht]
\centering
\caption{Baseline residual risk across locations}
\begin{tabularx}{\linewidth}{@{} >{\hsize=1\hsize}X >{\hsize=1\hsize}X @{}} 
\toprule
\textbf{Location} & \textbf{Baseline Residual Risk} \\ 
\midrule
L1: Straight Road & 0.491 \\ 
L2: T-Section & 0.941 \\ 
L3: Straight Road & 0.123 \\ 
L4: Complex Intersection & 0.968 \\ 
\bottomrule
\end{tabularx}
\label{tab:baselinerr}

\end{table}

Fig. \ref{fig:mean_residual_risk} shows the behavior of the mean residual risk as a function of ego localization uncertainty for all selected locations.

\subsubsection{L1 \& L3 Straight Road} 
Both straight road scenarios show a generally increasing behavior in mean residual risk with growing uncertainty. However, the response is not monotonic, which suggests that the belief distribution interacts with a spatially non-uniform residual risk field containing both higher and lower neighboring risk.

\subsubsection{L2 T-Section}
This scenario starts with an already high baseline residual risk. Under increasing localization uncertainty, the mean residual risk remains at a similarly high level with small fluctuations. This behavior indicates that uncertainty does not substantially increase risk further but distributes probability mass within an already high risk region. 

\subsubsection{L4 Complex Intersection}
The deterministic baseline of L4 is high, but small localization uncertainties slightly reduces the mean residual risk before it stabilizes at high levels for larger uncertainties. This indicates that the nominal ego position corresponds to a high risk region. Small uncertainties distribute the ego position into slightly less critical neighboring configurations.

\subsection{Residual Risk Variability}

The standard deviation of the residual risk over the Monte Carlo samples is visualized in Fig. \ref{fig:risk_variability}. For the straight road scenarios with lower baseline risk, variability increases with localization uncertainty, reflecting the broader spread of sampled ego poses across different risk regions. In contrast, for the T-Section and intersection scenarios with higher baseline risk, variation remains elevated across all uncertainty levels. Again, scenarios with already high deterministic residual risk tend to maintain high variability under uncertainty.  

\subsection{Exceedance Probability of Residual Risk}
\begin{figure}[!t]
    \centering
    \includegraphics[width=\linewidth]{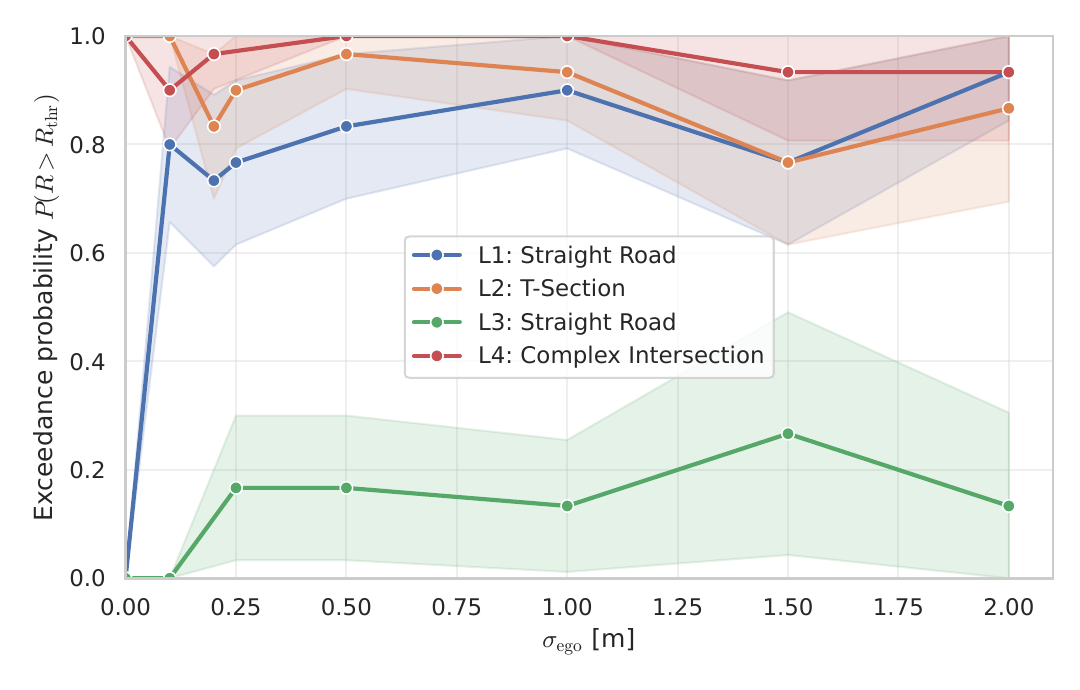}
    \caption{Exceedance Probability over increasing localization uncertainties for the selected scenarios}
    \label{fig:exceedance_probability}
\end{figure}

To quantify how often the system is pushed into high residual risk, the exceedance probability is defined as 
\begin{equation}
    P(R>R_{thr}) = \mathbb{P}(R(\mathbf{x}_{e})>R_{thr})
\end{equation}
where $R_{thr}$ depicts a predefined risk threshold, which is shown in Fig. \ref{fig:exceedance_probability}. For the straight road scenarios (L1 \& L3), the exceedance probability increases with localization uncertainty, reflecting the growing overlap of the ego belief distribution with high risk areas. In contrast, the T-Section (L2) exhibits an increase toward a high plateau, indicating thath even small localization uncertainties substantially raise the likelihood of critical configurations. The complex intersection (L4) maintains consistently high exceedance probabilities across all uncertainty levels. Overall, these results confirm that localization uncertainty both alters the expected residual risk but also adapts how frequently the system operates in high risk regions. This effect is scenario-dependent and linked to the spatial topology of the residual risk field. 

\subsection{Discussion}
The experiments show that ego localization uncertainty significantly influences residual risk estimation. However, this influence is scenario-dependent and governed by the spatial structure of the underlying residual risk field. Depending on the local road geometry at the nominal ego position, localization uncertainty may amplify, redistribute, or smooth the expected residual risk.

These findings demonstrate that residual risk cannot not be treated as a deterministic scalar quantity. Instead, it must be considered as a probabilistic measure defined over the ego position uncertainty. In geometrically complex environments, the nominal ego pose may already coincide with locally elevated risk regions, leading to high baseline risk even without additional uncertainty. Thus, localization uncertainty does not uniformly increase residual risk. Rather, its effect depends on the spatial configuration of the scenario.

From an AV system perspective, these results highlight the necessity of explicitly propagating localization uncertainty into risk assessment and planning modules. Although object detection often operates in the local sensor frame, tracking and prediction modules typically rely on globally referenced states and are therefore directly affected by pose uncertainty. Neglecting localization uncertainty in risk-aware decision-making can thus lead to wrong estimation of safety margins.

While the presented analysis provides an initial residual risk formulation incorporating localization uncertainty, some limitations must be acknowledged. On the one hand, the Monte Carlo estimation based on a finite sample size, which may limit the statistical stability of variability and exceedance estimates at higher uncertainty levels. Furthermore, the experiments are conducted under fixed velocity and prediction horizon setting, which restrict the generality of the findings to similar operating conditions. Future work should investigate these aspects to further generalize the proposed belief-space residual risk formulation.

%% file: sections/06_Conclusion.tex
\section{\uppercase{Conclusion}}
\label{sec:conclusion}

In this work, a residual risk concept is extended into belief space by introducing ego localization uncertainty into the risk computation. To analyze the effects of localization uncertainty on residual risk, a systematic analysis was conducted evaluating on a real world traffic dataset. The results show that localization uncertainty does not uniformly increase collision risk. Instead, its impact is dependent on the spatial topology of the residual risk field at the nominal ego pose. The analysis further present that spatial and temporal degradation mechanisms influence residual risk differently. While temporal latency smooth shifts in risk levels, localization uncertainty perturbs the geometric alignment between the ego vehicle and conflict structures. This means, that is has a scenario-dependent effect. Overall, the analysis highlights ego localization uncertainty as a crucial factor in risk assessment for AVs.